# A novel integrated method of detection-grasping for specific object based on the box coordinate matching

Zongmin Liu [1,2], Jirui Wang [2], Jie Li [2], Zufeng Li [1], Kai Ren [3,4], Peng Shi [5*]

1   National Research Base of Intelligent Manufacturing, Chongqing Technology and Business University, Chongqing 40006, PR China.
2   School of Mechanical Engineering, Chongqing Technology and Business University, Chongqing 400067, PR China.
3   State Key Laboratory of Fluid Power and Mechatronic Systems, school of Mechanical Engineering, Zhejiang University, Hangzhou, Zhejiang 310027, PR China
4   Key Laboratory of Advanced Manufacturing Technology of Zhejiang Province, school of Mechanical Engineering, Zhejiang University, Hangzhou, Zhejiang 310027, PR China
5   School of Electrical and Mechanical Engineering, University of Adelaide, Adelaide, SA 5005, Australia.
*   Correspondence: peng.shi@adelaide.edu.au

**Abstract:** To better care for the elderly and disabled, it is essential for service robots to have an effective fusion method of object detection and grasp estimation. However, limited research has been observed on the combination of object detection and grasp estimation. To overcome this technical difficulty, a novel integrated method of detection-grasping for specific object based on the box coordinate matching is proposed in this paper. Firstly, the SOLOv2 instance segmentation model is improved by adding channel attention module (CAM) and spatial attention module (SAM). Then, the atrous spatial pyramid pooling (ASPP) and CAM are added to the generative residual convolutional neural network (GR-CNN) model to optimize grasp estimation. Furthermore, a detection-grasping integrated algorithm based on box coordinate matching (DG-BCM) is proposed to obtain the fusion model of object detection and grasp estimation. For verification, experiments on object detection and grasp estimation are conducted separately to verify the superiority of improved models. Additionally, grasping tasks for several specific objects are implemented on a simulation platform, demonstrating the feasibility and effectiveness of DG-BCM algorithm proposed in this paper.

**Keywords:** Service robots; Object detection; Grasp estimation; Detection-grasping integrated algorithm

# 1. Introduction

After decades of development, there are now various types of service robots with different functions. However, it's still challenging to apply them in home scenes. It mainly due to the incomplete technologies related to object detection, grasp estimation, navigation, and positioning. With the backdrop of accelerated aging around the globe and the increasing demand for service robots in society, research and development in the field of home service robots have become an inevitable trend (Hsu et al., 2020). In order to provide care for disabled and elderly people, service robots combine with various technologies to accurately identify (Liu et al., 2022; Li et al., 2022) and track (Li et al., 2020; Li et al., 2020; Cao et al., 2022) targets. Therefore, the research on object detection and grasp technology of service robot in the indoor scenes holds significance theoretical and practical importance.

Before the emergence of deep learning, research on object detection and grasp technology primarily relied on traditional algorithms. In recognition and detection, the scale invariant feature transform (SIFT) algorithm (Lowe, 2004) was commonly used to identify key points and extract feature descriptors from images. (Dalal et al., 2005) created the histogram of oriented gradients (HOG) method that proved effective in detecting pedestrians. Regarding grasp technology, (Li et al., 2015) extracted 2D image edge features through SIFT and combined them with a depth camera to obtain 3D grasp pose estimation through iterative closest point algorithm. In contrast (Fan et al., 2018) introduced an adaptive grasp strategy that estimate the potential range of pose error for objects and compensate through the contact between fingers and objects during grasping. Furthermore, (Matsuda et al., 2022) combined image processing and distance measurement to grasp the target object.

The research mentioned above primarily relies on template feature matching and iterative calculation methods, which can have certain limitations. These approaches often involve large computational consumption and poor efficiency, leading to the difficulty of adapting to the unstructured environment with multiple confounding factors. Usually, service robots need to operate in dynamic and real-time situations.

This requires service robots to have the ability to perceive their environment, recognize objects, and interact with them through learning and reasoning processes. Traditional algorithms may struggle to meet these functional requirements due to their limited adaptability and scalability (Lee, 2019). With the deep learning developing rapidly, researchers have trained different convolutional neural networks (CNN) to obtain impressive feature extraction and expression capabilities in the field of computer vision. In this way, the CNN has achieved better results than traditional algorithms in object detection and grasp estimation that made their applications more extensive (Liu et al., 2021). For example, (Shang et al., 2020) proposed grasping posture prediction networks with CNN for dexterous robotic hand to predict optimal grasp postures and experimented in simulation. (Yu et al., 2021) used SSD and spatiotemporal long short-term memory (ST-LSTM) for space human-robot interactions in real-time hand gesture detection and identity-awareness. (Xie et al., 2021) combined deep convolutional generative adversarial network (DCGAN) and YOLOv4 to improve the accuracy of defect detection. (Qu et al., 2022) proposed a minimum bounding rectangle detection algorithm (MinBRect), which combined RGB and depth maps and calculated the minimum axis of the rectangle to quickly estimate grasp positions.

However, the above research only focused on improving grasping or detection separately while did not consider the detection and grasp process as a merged task, resulting in certain difficulty for robots to accomplish recognition and grasp. Furthermore, (Zhang et al., 2018) proposed the ROI-GD algorithm for grasp estimation in the ROI region obtained from object detection. However, this method performed well in stacked scenarios, it heavily relied on the accuracy of the ROI. (Li et al., 2023) proposed a two-step cascade system that combines YOLOv4 with GGCNN, which can simultaneously perform object detection and grasp estimation, but there was still room for improvement to achieve higher accuracy.

In summary, the current research on object detection and grasp estimation is relatively fragmented, resulting huge difficulties for robots to seamlessly integrate these tasks. In addition, due to the significant variations among different indoor scenes and the presence of interference such as occlusion, the degree of difficulty in precisely detecting and grasping objects is further increased. To enhance the capabilities of object detection and grasping of service robots and facilitate their widespread application, this paper proposed an integrated solution based on box coordinate matching for object detection and grasp estimation. The main contributions of our work are as follows

(1) An improved instance segmentation model based on the combination of convolutional neural networks (CNN) and attention mechanism was proposed as the branch for object detection.

(2) An improved planar grasp estimation model based on the ASPP and channel attention module that enhance the receptive field and feature extraction ability was built as the branch for grasp estimation.

(3) The DG-BCM algorithm was proposed to achieve specific object grasp. Furthermore, we built a robotic arm platform in simulation software and took experiments to verify the feasibility and effectiveness of improved models and the proposed algorithm.

The content of sections are arranged as follows: In Section 2 we summarized works related to this paper. In Section 3 we introduced the improved models and proposed algorithm of this paper while Section 4 described the experiments and results of above researches. Finally, we concluded the contribution of this paper and discussed the future direction in Section 5.

## 2. Relative work

### 2.1 Object Detection

Object detection algorithms based on deep learning are mainly include two methods, i.e.One-stage model and Two-stage model. One-stage models usually have a competent inference speed, such as YOLOv4 (Bochkovskiy et al., 2020), EfficientDet (Tan et al., 2019), etc. Two-stage models mainly use region proposal algorithm with a higher accuracy while the inference speed is slower, such as Faster RCNN (Ren et al., 2015), Cascade RCNN (Cai et al., 2017), etc. With the emergence of fully convolutional networks (FCN), researchers have proposed pixel level tasks such as semantic segmentation and instance segmentation. For example, (He et al., 2017) created Mask RCNN that added FCN branches to Faster RCNN. (Bolya et al., 2019) used the similar method on YOLO and proposed YOLACT. However, the above networks are all anchor based methods. (Wang et al., 2021) proposed the SOLO network based on anchor free that without the generation of bounding boxes. Instead, pixels in each target were directly classified based on their position and size that transform the detection and segmentation into classification.

On the basis, SOLOv2 (Wang et al., 2020) added dynamic convolution that can change the kernel weights through training and learning, and proposed Matrix NMS to get faster speed. The backbone of SOLOv2 is Resnet50 with five residual layers for feature extraction, while the neck takes FPN (Feature Pyramid Network) as the features fusion block for the output of different layers that can improve the model's discriminability for scales. The structure is shown in Fig. 1.

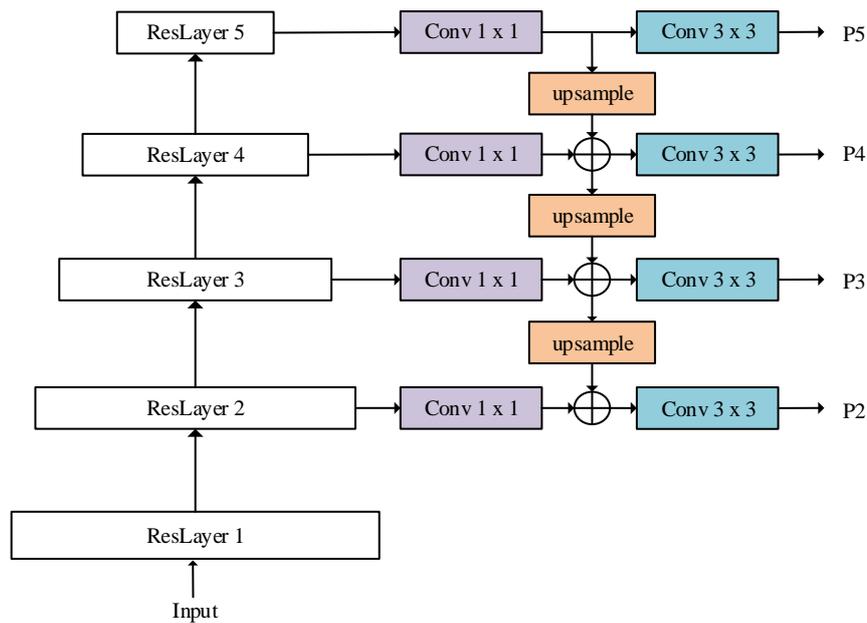

Fig.1. The structure of backbone and FPN

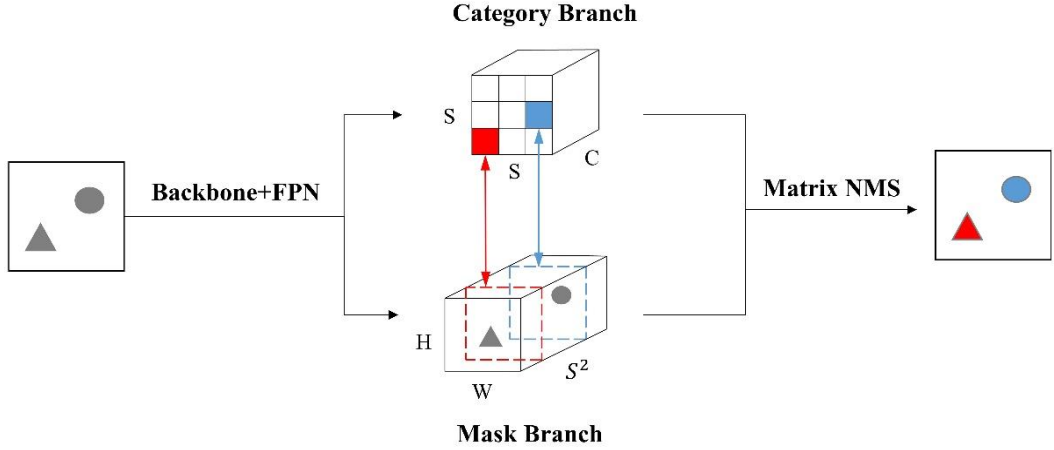

Fig. 2. The structure of SOLOv2 head

In addition, the category branch ($C \times S \times S$) and mask branch ($S^2 \times H \times W$) is included in the head, and the category branch divides the fused features into a grid of $S \times S$, mapped to the channels with $S^2$ in the mask branch, where each channel represents a predicted instance of grid, as shown in Fig. 2. If the center of an instance is located at the grid $(i,j)$ in category branch, the corresponding channel $k$ of mask branch is responsible for predicting the mask of the instance, which can be expressed as:

$$k = (i-1)*S + j \quad (i,j = 1,2,3,\dots,S) \tag{1}$$

where $i$ and $j$ represent the grid indices in the y and x directions, $k$ is the channel corresponding to mask branch, and $S$ is the number of grids in each direction. The Loss function of network can be composed of semantic category and mask prediction. The loss of mask prediction is defined as:

$$L_{mask} = \frac{1}{N_{pos}} \sum_k \alpha_{\{p^*_{i,j}>0\}} \cdot d_{mask}(m_k, m^*_k) \tag{2}$$

where $i,j,k$ is the same as (1), $N_{pos}$ is the total positive samples, $p^*$ and $m^*$ are the targets of category and mask respectively. $\alpha$ will be 1 if $p^*_{i,j} > 0$. Otherwise it set to 0. $d_{mask}$ is the Dice Loss (Milletarì et al., 2016). Then the total loss of training is defined as:

$$L = L_{cate} + \lambda L_{mask} \tag{3}$$

where $L_{cate}$ is the loss of semantic category that used the conventional Focal Loss (Lin et al., 2017), $\lambda$ is set to 3.

What's more, mask branch is further divided into two parts: mask kernel branch and mask feature branch. Mask kernel branch is for dynamical learning the weights of the kernel G for each layer of features output by FPN. After unifying the feature size of each layer to 1/4 of the original image and fusing them, input to the mask feature branch for convolution with G to predict the mask, and then obtain the corresponding instance mask through Matrix NMS.

## 2.2 Grasp Detection

The grasp pose estimation based on deep learning can be categorized into planar grasp and 6-DOF grasps, and this paper is focused on planar grasp. (Lenz et al., 2013) made a pioneering contribution by using neural networks to extract grasping features, laying the foundation for grasping estimation by deep learning. (Mahler et al., 2017) established the grasping dataset Dex-Net 2.0 and proposed a Two-stage detection algorithm of Grasp Quality Convolutional Neural Network (GQ-CNN), which first sampled candidate grasping poses and then evaluated the quality. However, this approach suffers from limitations such as high inference latency and poor performance in stacking scenarios of multiple objects. (Morrison et al., 2018) proposed a lightweight generative grasp convolutional neural network (GG-CNN), which only used depth images to achieve end-to-end grasp pose of each pixel, but the accuracy could be further improved. (Kumra et al., 2019) introduced the GR-CNN, and improve the ability of feature extraction through the residual module. Fig. 3 illustrates the structure of the network.

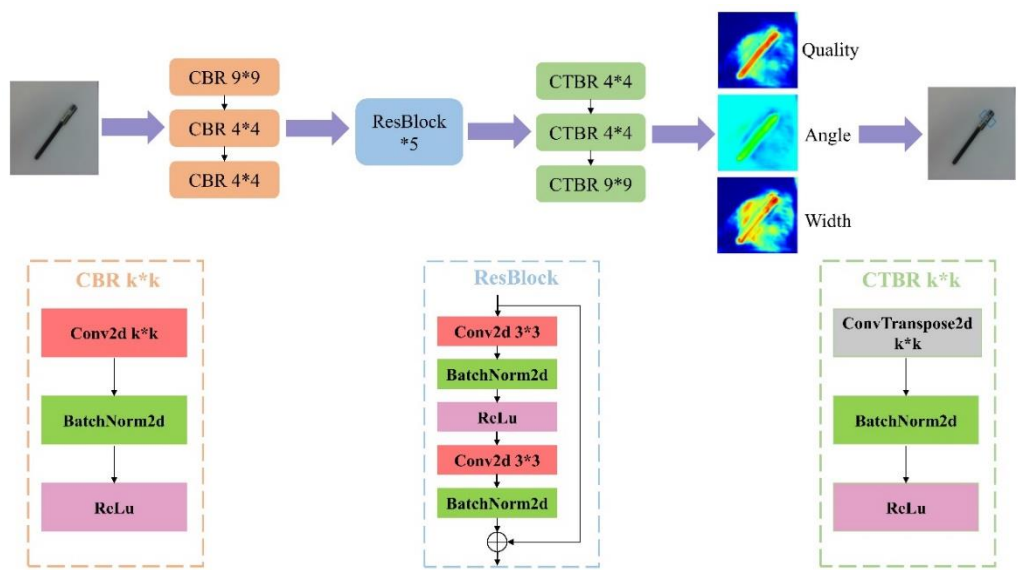

Fig. 3. The structure of GR-CNN

The network consists of CBR module, residual module and CTBR module. The modules mainly include convolution, batch normalization, activation, transpose convolution and others. For the input image, three CBR modules are used for pretreatment, and five residual modules for feature extraction. Finally, to preserve the spatial features of input data, up sampling is performed by transposing convolution, and get the grad-cam of Quality, Angle, and Width. The Quality is used to describe the grasp quality of pixels, and its value represents the grasp confidence score at all of pixels in image. The range is from 0 to 1, and if the score is higher, the success rate of grasp objects is higher. The Angle represents the grasp angle at all of pixels in image, ranging from $[-\pi/2, \pi/2]$. The Width is the width of grasp at all of pixels in image, ranging from $[0, W_{max}]$, where $W_{max}$ is the maximum width that the gripper can reach. Based on this, the definition of grasping posture is as follows:

$$g_i = (x_i, y_i, \theta_i, w_i, q_i) \qquad (4)$$

where $(x_i, y_i)$ is the center coordinate of the grasping rectangle. $\theta_i$ is the rotation relative to the frame of camera. $w_i$ is the width in pixel. $q_i$ is the quality score of $g_i$, and the model will output the highest quality score of grasp pose in priority.

At the same time, to convert pixel coordinates of the grasping pose into the coordinates of robot, the following transformation relationship can be used:

$$g_r = T_{rc}(T_{ci}(g_i)) \qquad (5)$$

where $T_{ci}$ is the camera intrinsics which is used to convert pixel coordinates into spatial coordinates under the camera frame. $T_{rc}$ is the eye-hand calibration matrix, which can transfer the coordinates from camera frame to robot frame to achieve grasping.

In loss function, for solving the exploding gradients, GR-CNN used the smooth L1 loss, defined as:

$$L(g_i, \hat{g}_i) = \frac{1}{n}\sum_{i=1}^{n} z_i \qquad (6)$$

$z_i$ is defined as:

$$z_i = \begin{cases} 0.5(g_i - \hat{g}_i)^2, & if \ |g_i - \hat{g}_i| < 1 \\ |g_i - \hat{g}_i| - 0.5, & otherwise \end{cases} \qquad (7)$$

where $g_i$ is the grasp pose predicted by network, $\hat{g}_i$ is the ground truth of grasp pose.

## 3. Method

Due to the variety of common objects in daily life, the accuracy of current target detection and grasp estimation models still have room for improvement in terms of accuracy. Therefore, we proposed the improved SOLOv2 and the improved GR-CNN respectively to promote the accuracy of object detection and grasp estimation. In addition, the DG-BCM algorithm that combined object detection and grasp estimation was introduced to complete grasping the specific objects.

**3.1 The improved SOLOv2**

The original SOLOv2 used Resnet50 as the backbone and FPN as the neck. Inspired by CBAM (Woo et al., 2018), we split it into channel attention module (CAM) and spatial attention module (SAM), and inserted into the SOLOv2. The structure of CAM is shown in Fig. 4. For the $F$, it will be processed by global maximum pooling (GMP), global average pooling (GAP) and shared multi-layer perceptron (MLP). Then, the corresponding elements are added and processed by sigmoid. Finally, the result is multiplied with $F$ in channel to get $F_c$. This process can be defined as:

$$F_c = F \otimes \sigma(MLP(GAP(F)) \oplus MLP(GMP(F))) \qquad (8)$$

where $\oplus$ is the addition of matrix on corresponding elements. $\otimes$ is the multiplication of matrix on corresponding elements, $\sigma$ is the sigmoid function, defined as:

$$\sigma(x) = \frac{e^x}{e^x + 1} \qquad (9)$$

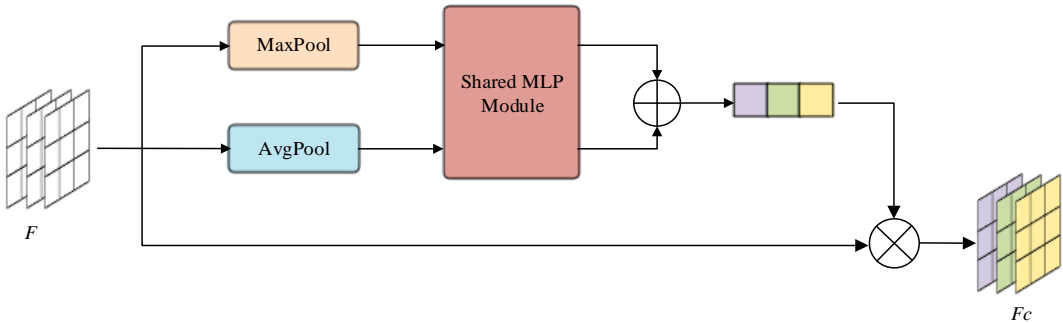

Fig. 4 The structure of CAM

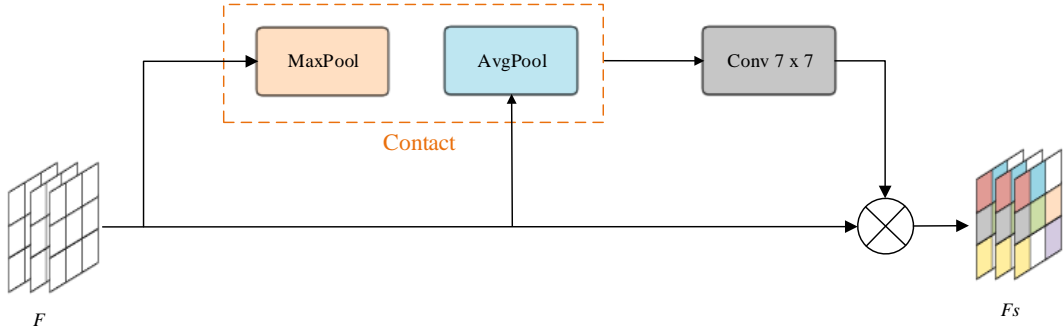

Fig. 5. The structure of SAM

The structure of SAM is shown in Fig. 5. GMP and GAP are performed based on channel respectively. The obtained results are contacted, then subjected to 7 x 7 convolution and sigmoid, and multiplied by the corresponding elements of matrix to obtain $F_s$:

$$F_s = F \otimes \sigma(Conv_{7\times 7}([GAP(F), GMP(F)])) \quad (10)$$

where $Conv_{7\times 7}$ represents a 7x7 convolution.

We place SAM at the end of ResLayer 1-4 in the backbone to improve the extraction of spatial feature details. CAM was placed at the end of each leave in FPN to optimize the different channel weights of fused features. The improved structure is shown in Fig. 6.

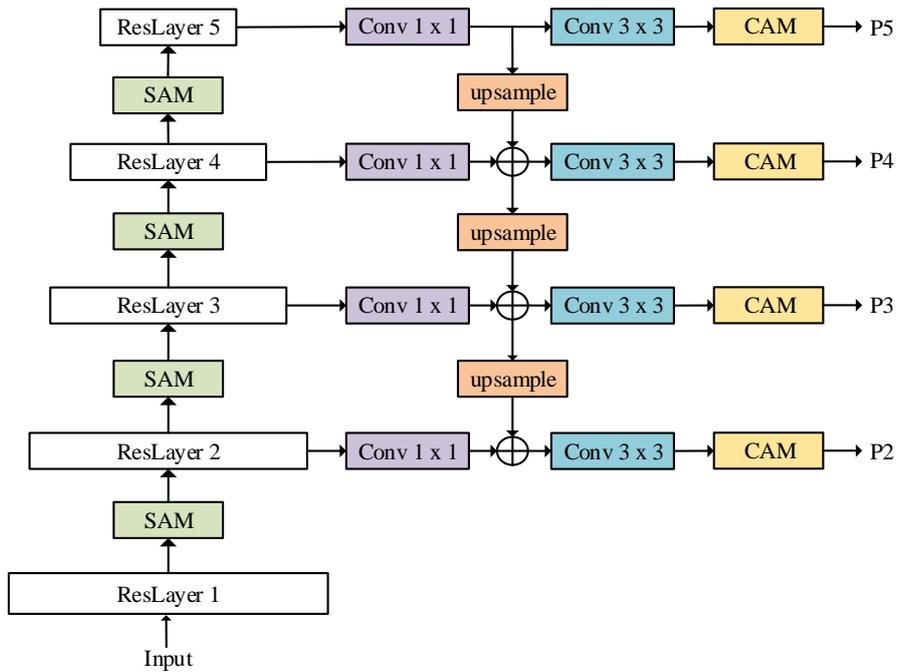

Fig. 6. The structure of improved model

## 3.2 The improved GR-CNN

In order to improve the feature extraction quality of GR-CNN, we added the ASPP (Chen et al., 2016) to the network. The core idea of the module is the same as the spatial pyramid pooling (SPP), while the improvement is to expand the receptive field of convolution kernels without decreasing the quality of resolution by introducing atrous convolution instead of down sampling. The network structure is shown in Fig. 7.

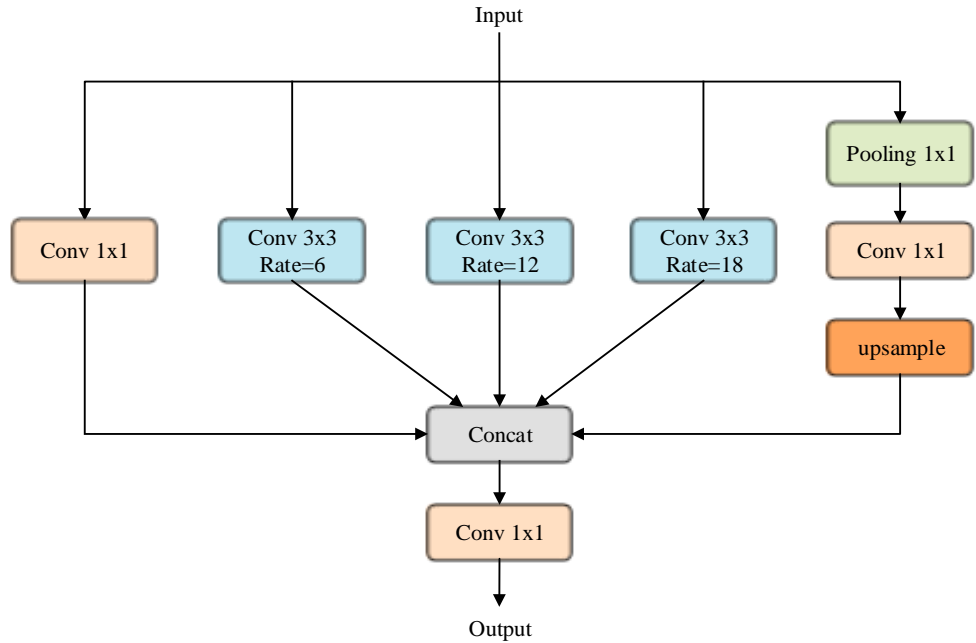

Fig. 7. The structure of ASPP

ASPP consists of a 1x1 convolution, a pyramid pooling that include 3x3 atrous convolutions with different dilation rates and a 1x1 adaptive average pooling. The dilation rates of each branch in the pyramid pooling can be customized to achieve multiscale feature extraction to expand the receptive field. This module was placed behind CBR in GR-CNN. In addition, considering that the trained Cornell dataset has almost white backgrounds that contrast significantly with the objects, we added CAM into residual module to extract meaningful information by emphasizing color-related channel features. The improved GR-CNN structure is shown in Fig. 8.

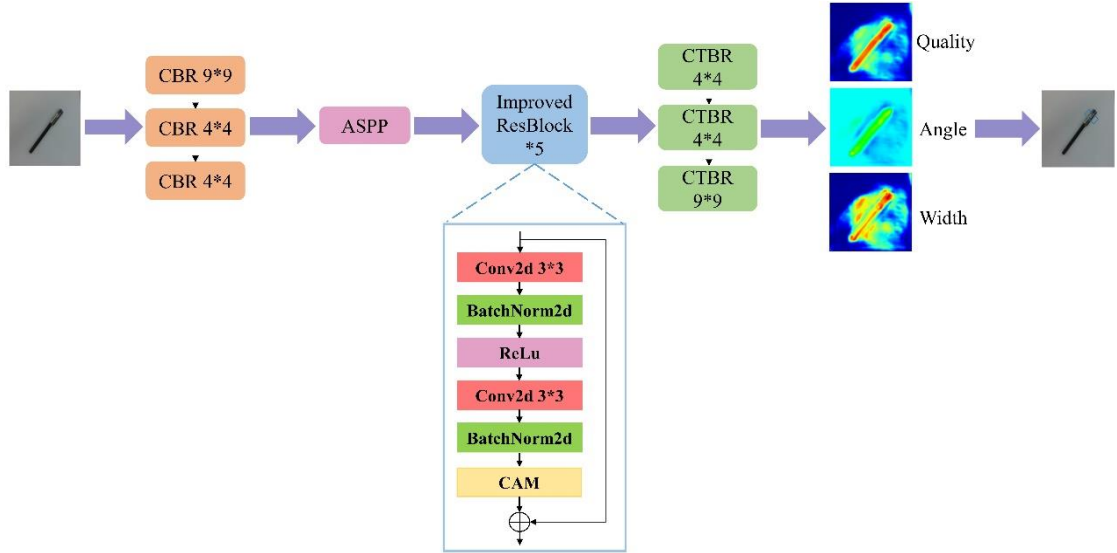

Fig. 8. The structure of improved GR-CNN

**3.3 DG-BCM algorithm**

In this section we illustrated the DG-BCM algorithm, and the improved SOLOv2 and GR-CNN were used to get the necessary information of targets about labels, bounding box coordinates and grasp poses. The framework of the algorithm is shown in Fig. 9.

Firstly, an image is processed with improved SOLOv2 and GR-CNN. In SOLOv2 branch, detected labels of objects and corresponding box coordinates of top left and bottom right are obtained. Then we store the corresponding label and coordinates information in a dictionary. In this dictionary, the labels serve as the keys, and the corresponding coordinates are stored as the values. According to the specific class, the algorithm will query all keys in the dictionary and obtain the correct value. In the meanwhile, a number of grasp rectangles with center coordinates are obtained in GR-CNN branch. By determining whether the center coordinates of rectangle are within the detection box, the grasp pose for the specific target is singled out. Due to the fact that the grasp pose is arranged in accordance with quality scores in descending order, our algorithm will first get the rectangle with the highest score and discard the others when there are two or more rectangles for an object like the handset in Fig. 9.

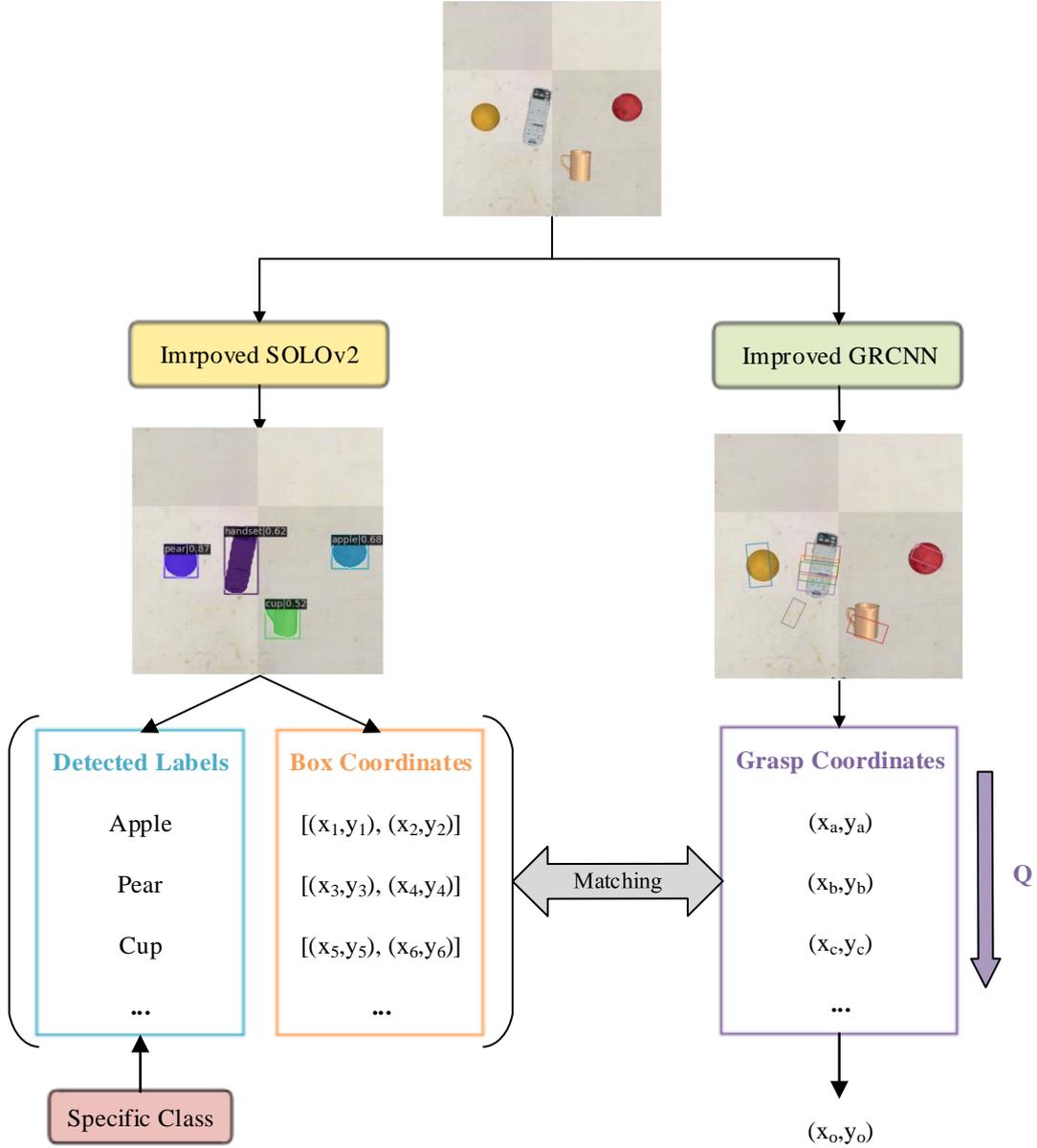

Fig. 9. The architecture of DG-BCM algorithm

This process can be defined as follows:

$$g_o = Q_{max}(g_1, g_2, \ldots, g_n) \tag{11}$$

Where $Q_{max}$ represents to get the maximum quality score. $g_1, g_2, \ldots, g_n$ are the grasp poses located in specific target. $g_o$ is the final output of grasp pose.

## 4. Experiments and results

### 4.1 Platform

The experiment was conducted on a Win10 PC with the following hardware configuration: an Intel Core i7-10700 CPU @ 2.90GHz, 16 cores, a GeForce GTX 1650 GPU with 4GB VRAM, and 16GB of memory. PyTorch was chosen as the environment for the model training and inference. CoppeliaSim was used as the simulation software and a Kinova robotic arm grasping platform was created. By combining remote control API with Python, the simulation and deep learning were combined to complete the visual grasping task.

The detected and grasped objects in the experiment include five categories: cola (cylinder-shaped), cup (cylinder-shaped with a handle), handset (cuboid), apple and pear (spheres with different colors). These items are very broadly representative in indoor living scenarios. Since there were no 3D models of cola and handset, we used Colmap to perform three-dimensional reconstruction that mainly includes four steps: data collection, feature matching, sparse reconstruction, and dense reconstruction. After post-processing by Meshlab, we completed the conversion from 2D images to 3D models. Five models are shown in Fig. 10.

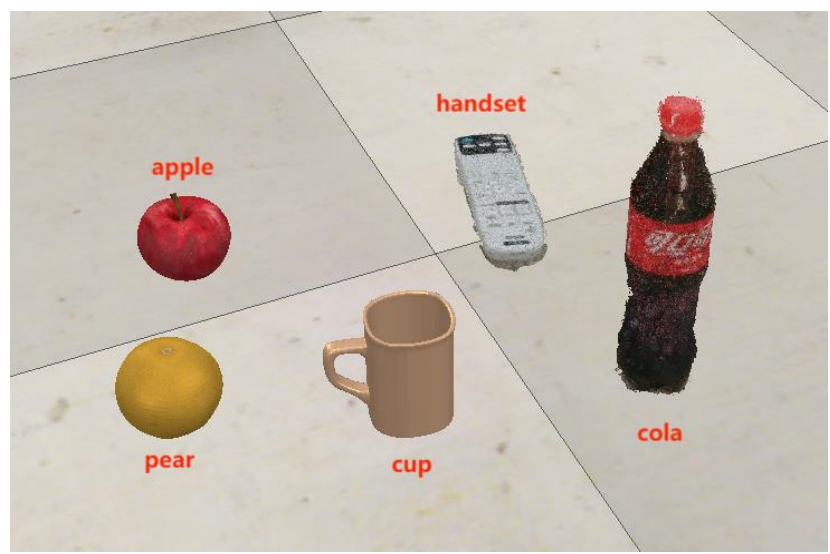

Fig. 10. The models of five targets

**4.2. Experiments**

We conducted experiments from three aspects about detecting and identifying targets based on improved SOLOv2, grasping estimation based on improved GR-CNN, and combining the two to achieve grasp of specific target in simulation environment.

**4.2.1 Object detection**

To compare the performance of the improved SOLOv2 with the original model, several steps were taken. Initially, we collected and annotated data for the five targets mentioned above, and converts them to the COCO (Lin et al., 2014) dataset format. In training, the SGD optimization algorithm was considered, and epoch was 20 due to taking the pre-trained weights for transfer learning. Batch size was2. The initial learning rate was 0.001, and the single cycle cosine decay strategy was adopted. The decay ratio was 0.1, and momentum was 0.9. The training loss curve is shown in Fig. 11, where blue represents the loss of original model, and orange represents the loss of improved model. It can be found that the loss of improved model is more stable with similar loss values of original at the process of the training.

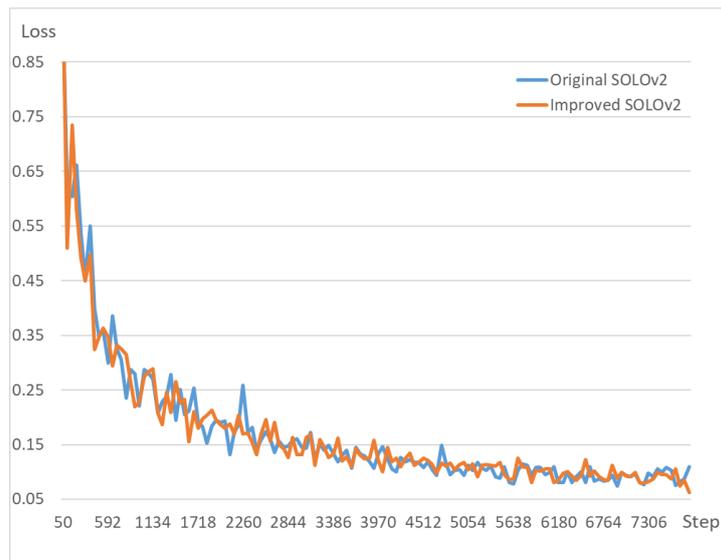

Fig. 11. The loss of original and improved SOLOv2

To verify the performance of the improved model with interferences of similar shape, different color and occlusion, apple, pear, and cola were considered to detect in the first experiment. The results are shown in Fig. 12.

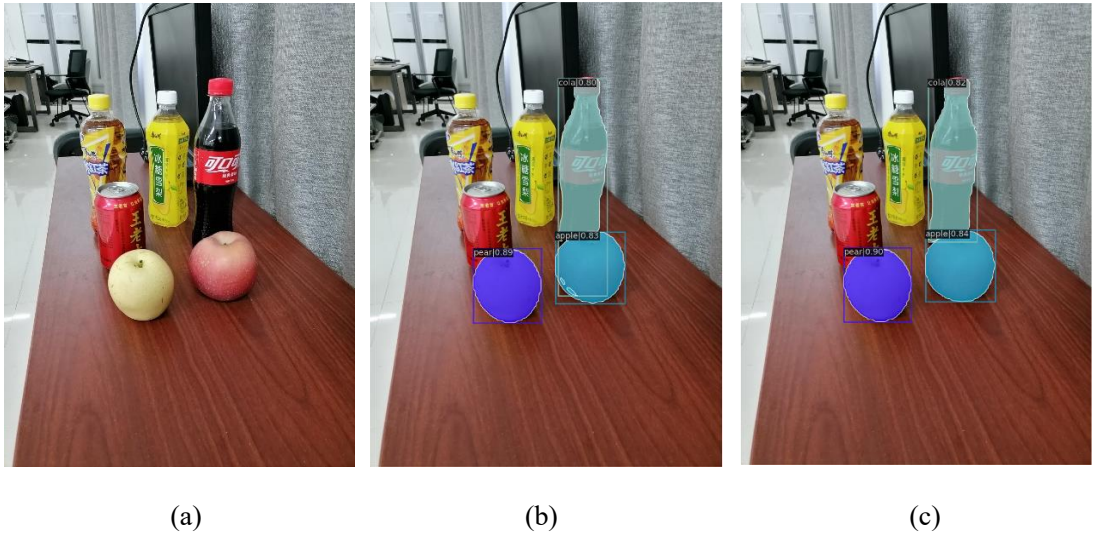

(a)                      (b)                      (c)

Fig. 12. The detection results of apple, pear, and cola: (a) Original image; (b) SOLOv2;

(c) Improved SOLOv2

Based on the information provided, it was evident that the improved SOLOv2 model exhibits higher confidence levels compared to the original model for all the targets. Additionally, in the overlapping region between cola and apple, the improved model demonstrates better recognition ability under different interferences.

The next experiment focuses on testing the performance of the improved model on cups and handset. Among them, there are two cups with different color and shape that require sufficient anti-interference ability. The results are presented in Fig. 13.

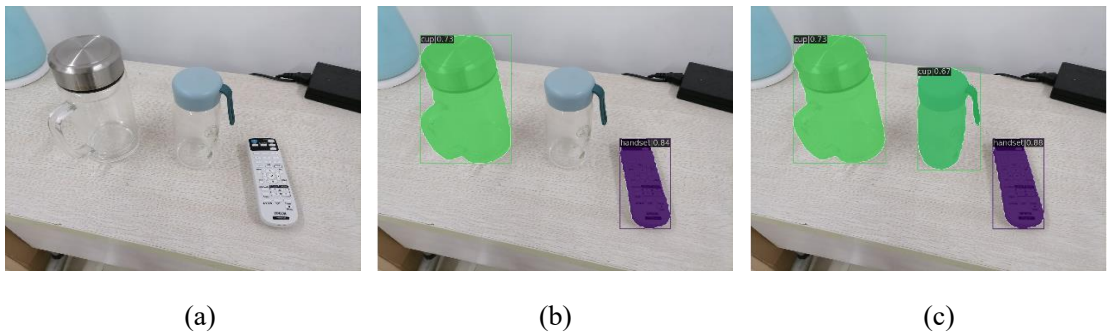

(a)                      (b)                      (c)

Fig. 13. The detection results of cup and handset: (a) Original image; (b) SOLOv2;

(c) Improved SOLOv2

It was apparent that due to the different colors and shapes of cups, the original model failed to detect the cup with a blue lid in the middle, while the improved model can effectively detect it. In terms of evaluation, we refered to the standard of COCO dataset and comprehensively measured the model performance through mAP (Mean

Average Precision). The results on the test dataset are shown in the Table 1. It is obvious that the improved model outperforms the original model in both $mAP^{0.5}$ and $mAP^{COCO}$.

Table 1 Evaluation Results of Detection Models

| Model | $mAP^{0.5}$ | $mAP^{0.75}$ | $mAP^{COCO}$ |
|---|---|---|---|
| SOLOv2 | 0.9415 | 0.9206 | 0.8321 |
| Improved SOLOv2 | 0.9517 | 0.9096 | 0.8479 |

**4.2.2 Grasp estimation**

Similarly, the original model was also the comparison. The Cornell grasp dataset was used for training and validation, and the superiority of the improved model was verified through training loss, accuracy, and experiment. In the training, the SGD was again utilized, and batch size was 4. Since the pre-trained weight was not used, the epoch was 80. The initial learning rate was 0.01. The decay strategy, ratio and other parameters were the same as object detection. Finally, the loss curves of the two models are shown in Fig. 14, where blue represents the original model and orange represents the improved model. It is evident that the improved model has less training loss.

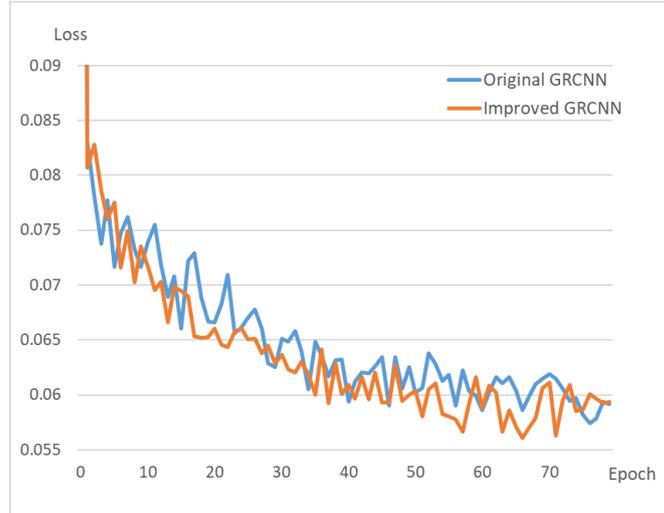

Fig. 14. The loss of original and improved GR-CNN

In terms of accuracy, we calculate the grasp success using the IoU (Jacquard) metric (Mahler et al., 2017). We assumed $G_{gt}$ as the ground truth of grasping rectangles, and $G_p$ as the grasping rectangle predicted by the model. If the angle between the two exceeds 30°, the predicted $G_p$ was considered incorrect and discarded. Otherwise, the IoU of $G_{gt}$ and $G_p$ can be defined as:

$$\text{IoU}(G_{gt}, G_p) = \frac{G_{gt} \cap G_p}{G_{gt} \cup G_p} \tag{12}$$

If the IoU is greater than 0.25, the predicted grasping rectangle is considered correct, or it is also discarded. Based on this method, we define the accuracy Acc as follows:

$$\text{Acc} = \frac{N_c}{N_{gt}} \tag{13}$$

where $N_{gt}$ is the number of ground truths, $N_c$ is the number of correctly predicted rectangles.

After every epoch of training, it is verified on the validation dataset to obtain the curve of accuracy, as shown in Fig. 15. Blue represent the original model while orange represent the improved model, and it can be seen that the improved model has higher prediction accuracy.

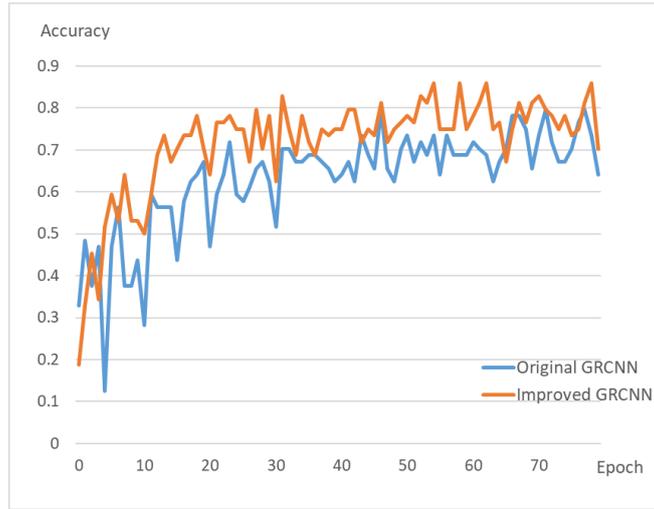

Fig. 15. The accuracy of original and improved GR-CNN

In addition, grasping pose estimation experiments of five targets were conducted in simulation, and the results are shown in Fig. 16.

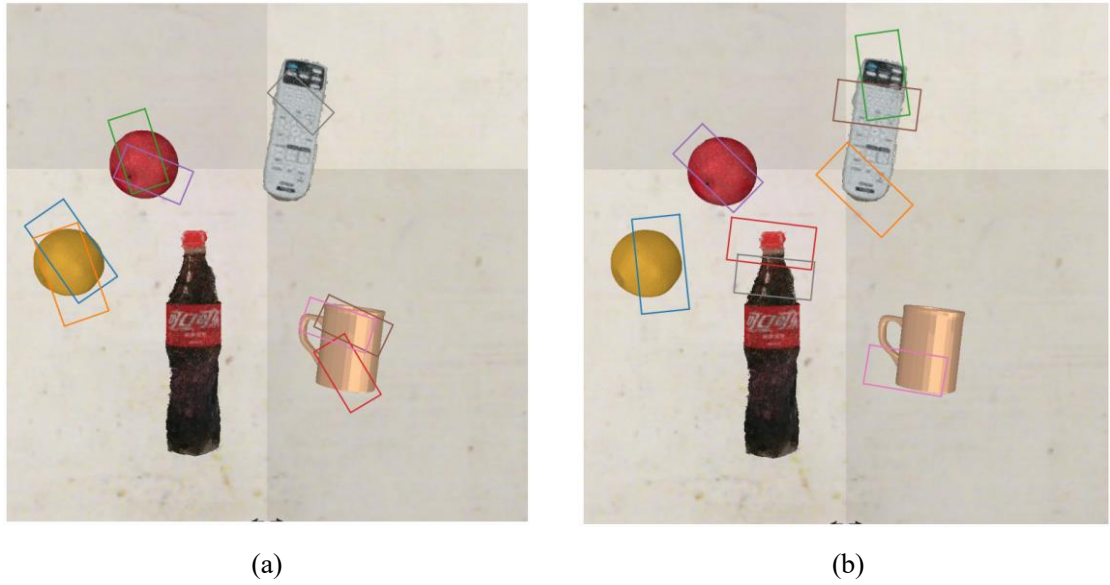

(a)                      (b)

Fig. 16. The grasp estimation results of five targets: (a) GR-CNN; (b) Improved GR-CNN

Obviously, the original model is unable to estimate the grasp pose of cola, and the performance of estimation on handset is poor. However, the improved model effectively overcomes the above problems, and the estimation of other targets remained good, making it more suitable for robot grasping experiments.

**4.2.3 Grasp simulation**

We conducted detection and grasping experiments for five targets in simulation. Firstly, the inference results from improved SOLOv2 and improved GR-CNN were obtained, as shown in Fig. 17. Secondly, based on the DG-BCM algorithm proposed in this paper, the specific object's detection box was obtained where the category name was entered by the user. In this way, the central coordinates of multiple grasp rectangles were matched with detection box by turn to obtain the unique optimal grasp pose. Thirdly, we transformed the grasp pose into robot coordinate frame to get the target position, and obtained the rotation of each joint on robotic arm through the Damped Least Square (DLS) inverse algorithm. Finally, the gripper will reach to the target position to complete the grasping task. The results of grasping five targets are shown in Fig. 18.

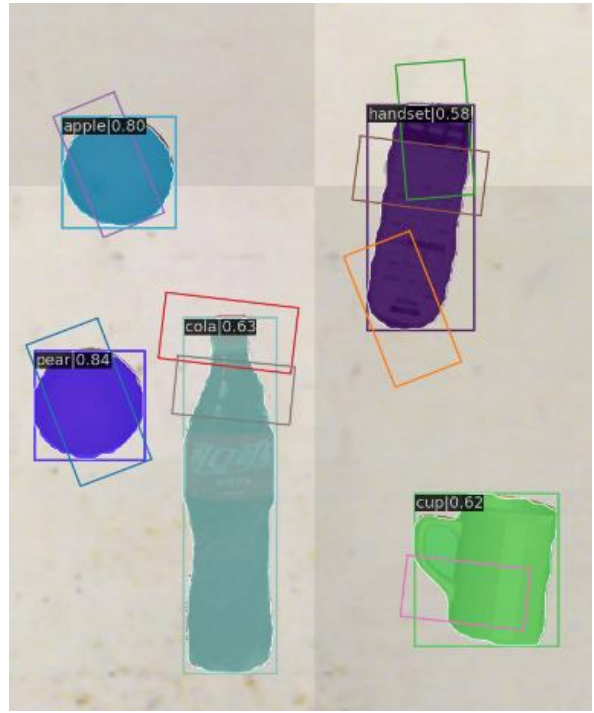

Fig. 17. The inference of improved SOLOv2 and improved GR-CNN

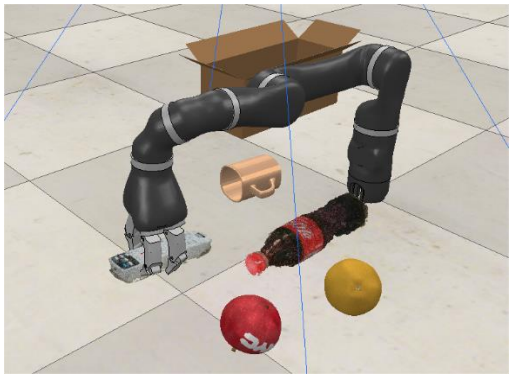

(a)

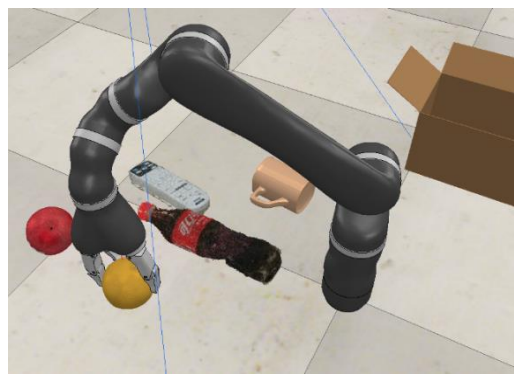

(b)

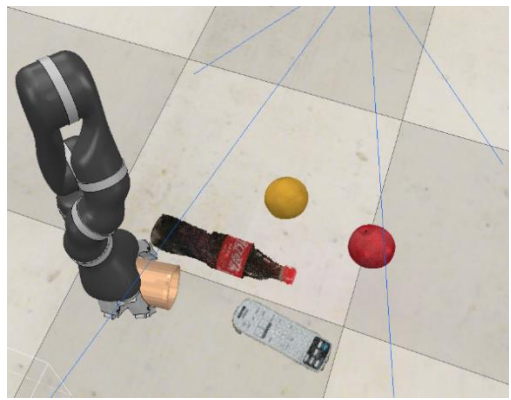

(c)

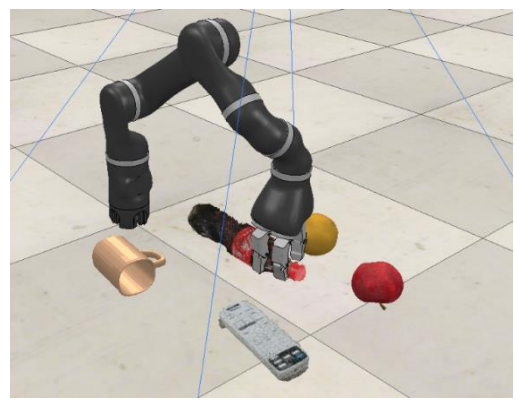

(d)

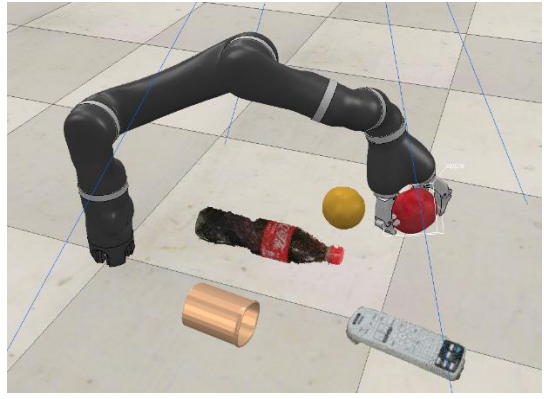

(e)

Fig. 18. The grasp results of five targets with DG-BCM algorithm: (a) Handset; (b) Pear; (c) Cup;

(d) Cola; (e) Apple

The rationality and feasibility of DG-BCM algorithm proposed in this paper have been verified through the above experiments.

**4.3 Discussion**

In this section, we individually adopted experiments for improved models of SOLOv2 and GR-CNN, and grasping specific object with DG-BCM method. The object detection results showed that improved SOLOv2 is able to accurately detect all targets with the higher values of $mAP^{0.5}$ and $mAP^{COCO}$. In addition, the loss values are basically the same, but with better stability. The grasp estimation results showed that improved GR-CNN has estimated better grasp poses of targets and got preferable loss and accuracy than original. The grasp simulation indicated that the DG-BCM has the practicability of grasping specific objects. From the above, it is apparent that our work presented in this paper is feasible to detect and grasp the objects according to user's requirement that could further promoted the application of service robots.

## 5. Conclusion

To improve the accuracy and efficiency of object detection and grasp estimation for service robots, as well as promote the collaborative development, we carried out researches from the following aspects.

Firstly, channel attention and spatial attention modules were added to SOLOv2 for object detection. The superiority of the improved model was verified by comparing mAP and experiments while the $mAP^{0.5}$ is 0.9517 and $mAP^{COCO}$ is 0.8479, higher than the 0.9415 and 0.8321 of original model.

Secondly, ASPP and channel attention module were added to the GR-CNN for grasping estimation, and the best accuracy of improved model is 0.859 in training that is higher than the 0.797 of original model.

Finally, we fused two improved models and proposed DG-BCM algorithm based on box coordinate matching. The experiments of detection and grasp for specific objects were completed via simulation experiments, validating the feasibility of the algorithm.

In the future, the number of recognition categories will be further expanded to enrich the objects that robots can recognize and grasp. What's more, we believe that the SOLOv2 can also be optimized in terms of lightweight, and there is a possibility of further improvement in the estimation accuracy of GR-CNN that we will take these aspects into consideration.

## CRediT authorship contribution statement

**Zongmin Liu:** Conceptualization, Supervision, Data curation, Writing– review & editing. **Jirui Wang:** Methodology, Writing – original draft. **Jie Li:** Validation, Experiment. **Zufeng Li:** Data curation. **Kai Ren:** Writing – review & editing. **Peng Shi:** Conceptualization, Supervision, Writing – review & editing.

## Declaration of competing interest

The authors declare that the research was conducted in the absence of any commercial or financial relationships that could be construed as a potential conflict of interest.

## Data availability

The data that support the findings of this study are available from the corresponding author upon reasonable request.

## Acknowledgments

This work was supported by grants of the National Key Research and Development

**References：**